# Multi-regime analysis for computer vision-based traffic surveillance using a change-point detection algorithm

SEUNGYUN JEONG, AND KEEMIN SOHN
Department of Urban Engineering, Chung-Ang University, Seoul 06974, South Korea

Corresponding author: Keemin Sohn (kmsohn@cau.ac.kr).

**ABSTRACT** As a result of significant advances in deep learning, computer vision technology has been widely adopted in the field of traffic surveillance. Nonetheless, it is difficult to find a universal model that can measure traffic parameters irrespective of ambient conditions such as times of the day, weather, or shadows. These conditions vary recurrently, but the exact points of change are inconsistent and unpredictable. Thus, the application of a multi-regime method would be problematic, even when separate sets of model parameters are prepared in advance. In the present study we devised a robust approach that facilitates multi-regime analysis. This approach employs an online parametric algorithm to determine the change-points for ambient conditions. An autoencoder was used to reduce the dimensions of input images, and reduced feature vectors were used to implement the online change-point algorithm. Seven separate periods were tagged with typical times in a given day. Multi-regime analysis was then performed so that the traffic density could be separately measured for each period. To train and test models for vehicle counting, 1,100 video images were randomly chosen for each period and labeled with traffic counts. The measurement accuracy of multi-regime analysis was much higher than that of an integrated model trained on all data.

## I. INTRODUCTION

Computer vision technology has been widely adopted to measure traffic parameters such as traffic volumes, densities and speeds [1-5]. This trend was accelerated when deep-learning models could recognize objects within an image similar to the way a human does [6-14]. In particular, vehicle detection is an easier task than recognizing other objects, because a vehicle takes a simple shape and the road environment for traffic surveillance is not complex. Many researchers are trying to devise a robust vehicle detector for the use of traffic surveillance based on deep learning [15,16]. For such a detector to be installed in the field for traffic surveillance, however, some limitations must be resolved. The current level of detection accuracy varies according to ambient conditions such as the time of day, weather and shadows. It remains difficult to develop a single vehicle detector with consistent performance both at daytime and nighttime. To recognize vehicles at nighttime, some researchers have utilized headlight beams, which is a feature that is unique to vehicles [17]. Switching detection models, however, cannot be automatically implemented on a real-time basis.

The present study provides a robust way to determine the switching times based on consecutive video frames under the assumption that the video frames reflect ambient conditions. Although it is easily understood that video images include hidden features of ambient conditions such as the time of day, how to extract them and to reduce their dimensions into a tractable level is a different story. We decided to use an autoencoder to elicit ambient conditions from video images. Whereas conventional principal component analysis (PCA) can deal with linear relationships when reducing dimensions of input variables in machine learning, an autoencoder can extract factors based on nonlinear correlations between variables. The hidden features an autoencoder extracts are then used as input for an algorithm to determine the change-points. A novel algorithm for finding change-points was adopted in the present study [18]. This algorithm depends on a rigorous statistical test unlike other rule-based naïve methods. This algorithm also offers the advantage of field application, because it recognizes the changes in ambient conditions on a real-time basis.

We set up a deep-learning-based regression model for vehicle counting to validate the multi-regime analysis. A day was divided into 7 different periods, each of which had unique ambient conditions, and a set of model parameters was separately trained for each period in advance. The same amount of image data was sampled for each period and used for training and testing tasks. As a reference, a model was trained and tested on the entire dataset for all periods under the expectation that the multi-regime analysis would outperform the integrated analysis.

The next section introduces both an autoencoder to abstract images and a change-point detection algorithm to separate multiple regimes. The third section describes how to choose a testbed and shows how the ground truth change-points were determined for the testbed. The fourth section lists the results of implementing a change-point detection algorithm when several different input dimensions are processed by an autoencoder. The section also verifies the validity of the multi-regime approach to count vehicles for traffic surveillance. The last section draws conclusions and suggests further studies.

## II. Modeling Framework

### A. The Autoencoder Reduces the Dimension of Images

A video image is represented by three values of RGB for all pixels that comprise it. If these values are used as an input variable to find change-points, the computation time increases exorbitantly. The most popular method to reduce the variable dimension is to use a PCA [19,20], but this algorithm has a handicap whereby only a linear relationship between variables can be accommodated when finding factors with a reduced dimension. On the other hand, an Autoencoder can consider non-linear correlations between original input features when reducing their dimension. For this reason, Autoencoders have been widely utilized in machine learning for dimension reduction [21,22]. The present study devised an Autoencoder to change a 128×512×3 mage into only 3 to 10 features. Fig. 1 shows the model architecture of the Autoencoder established in the present study. The encoding part is composed of 6 hidden layers. After flattening the last convolutional layers, 5 fully connected (FC) layers are attached to reduce the feature dimension. The number of nodes in the last FC layer of the encoding part decreases to the three target dimensions (3, 5 and 10). The last layer of the encoding part is amplified again to reproduce the original input dimension through the decoding part.

Training an Autoencoder requires no supervision with labeled data. The output image that an Autoencoder evaluates from an input image is refitted to the input image. Such a self-fitting algorithm guarantees obtainment of the reduced features in the last layer of the encoding part, and only node values of the middle layer of an Autoencoder are used as variables for the subsequent change-point detection algorithm once all parameters are completely trained. Variations in ambient conditions was expected to be determined by a change-point algorithm that used the middle node values elicited from a trained Autoencoder.

The video image of a testbed includes both ambient and traffic conditions. We adopted a two-step process to count vehicles in the testbed. In the first step, only the features for ambient conditions are extracted from images, and the times of day are divided based on these. In the second step, a regression-based model is set up to count vehicles in an image, and model parameters are separately trained and tested for each period. The features derived in the first step, however, should include both ambient and traffic conditions. To suppress the features of traffic conditions, we tried several different numbers (3, 5, and 10) of final features (= numbers of middle nodes in the proposed Autoencoder) and examined how each option performed in the counting of vehicles.

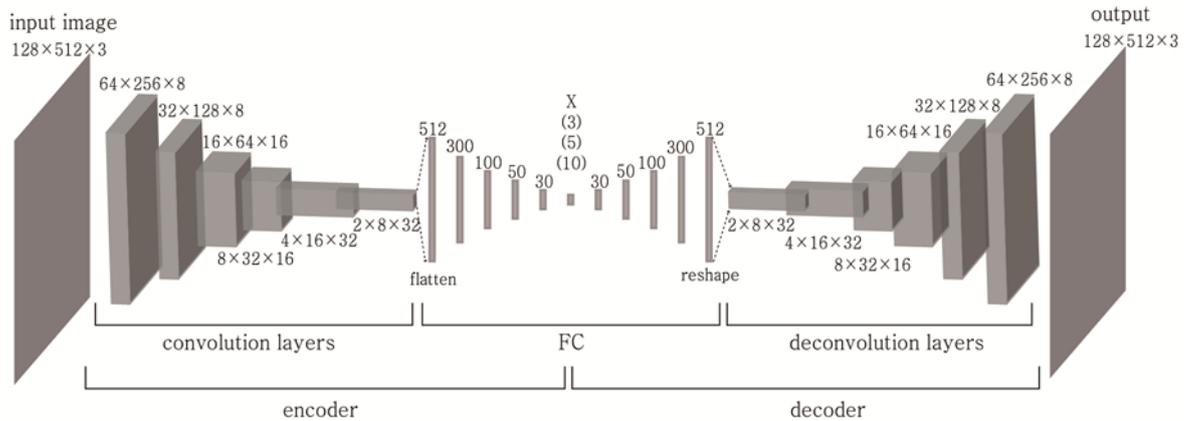

**FIGURE 1.** Architecture of the proposed autoencoder used to reduce a feature dimension

### B. Change-Point Detection Algorithm

Change-point detection algorithms are widely using in science and engineering fields. Applications include signal processing, dynamic social networks, and online marketing. In similar manner, various change-point detection algorithms have been developed such as the segment neighborhood model [23], piecewise IID data models [24], the structural change model [25], PELT [24], and the Group-Fused Lasso (GFL) [26]. Whereas a univariate model prevails in the field [24-26], few models can successfully accommodate multivariate problems [27-29]. Other researchers recently developed a robust multivariate algorithm under the assumption that target data follows a parametric probability density [18]. This assumption makes it possible to use a statistical test to judge the validity of

3 | 11

a change-point. The present study applied an algorithm that separates different regimes of ambient conditions in video images for traffic surveillance.

Once an image is represented in a smaller dimension after being processed by an Autoencoder, the abstracted features constitute multivariate time-series data. First, the data are regarded as random variables with a Gaussian probabilistic density. Parameters of the Gaussian distribution include the means, variances and covariances. The number of parameters to be estimated varies according to the feature dimension. Second, the abstracted features can be assumed to follow a Dirichlet probabilistic distribution. To use the Dirichlet distribution, the features must be compositional data that take a proportional value and sums up to 1. Such data encompasses the percentage contribution of goods sold every month, the proportion of time spent by each task that comprises a whole production process in a given time period, and probabilities from a discrete distribution, etc. Data must be preprocessed so that the Dirichlet probabilistic distribution can be applied. Raw data must be placed on a simplex in a hyperspace, and to do so the raw data of each interval should be changed to a compositional form via the use of a multi-dimensional Expit function (inverse of multinomial logit function), the details of which are described later in this section.

The theory behind the change-point detection algorithm in reference [18] is described as follows. The symbols $\{x_1, \ldots, x_T\}$ denote the sequence of feature variables where $x_i \in R^d$ is a feature vector at interval $i \in \{1, \ldots, T\}$, $T$ denotes the total number of time periods, and $d$ is the dimension of the feature vector that corresponds to the size of the middle layer of an Autoencoder. The change-point detection determines $\{\tau_1, \ldots, \tau_{k-1}\}$, where $k$ is the number of regimes to be differentiated and $\tau_i$ is the change-point that separates the $i^{th}$ period from the $(i+1)^{th}$ period ($i \in \{1, \ldots, k-1\}$). These multiple change-points are detected by repeatedly implementing a single change-point detection algorithm. Once an active window size (I) is determined, the portion of the sequence data that is covered by the window can then be investigated from the start point. The initial search period then becomes [1, I]. If a change-point is identified within this time window, it becomes the first change-point, $\tau_1$, and the next time range is moved to [$\tau_1$, $\tau_1$+I]. When finding a change-point within the current range fails, the next search range is increased to [1, $I + B$] by a predefined update interval ($B$). This process continues until the end-point of the incumbent range reaches the final interval, $T$. An online implementation of the process is possible once a small $B$ is selected.

A single-point detection algorithm is dependent on the assumption that variables at different intervals would be independently distributed according to a parametric probability density. We tested two parametric probability densities, a Gaussian distribution and a Dirichlet distribution, to test whether a change-point can divide a period into two different entities, and a null hypothesis ($H_0$) is set up whereby data within a range ($A = [1, t]$) come from a single probabilistic distribution. Under the null hypothesis, the maximum log-likelihood can be computed by Eq. (1) such that the probability that the data are collected could be maximized.

$$LL_0 = \max_\theta LL(\theta)$$
$$= \max_\theta [\log(f_\theta(x_1)) + \cdots + \log(f_\theta(x_t))] \qquad (1)$$

In Eq. (1), $f_\theta$ is a multivariate probability density function of $d$-dimensional feature variables, and $\theta$ is the vector of function parameters to be estimated.

The log-likelihood under the null hypothesis is a baseline that computes a test statistic to reject the hypothesis and to conclude that a change-point exists within the range. To compute the statistic, another log-likelihood is necessary. The log-likelihood value should be computed under the alternative hypothesis ($H_\tau$). The alternative hypothesis says that a given range would be divided into two sub-ranges, each of which has a different set of parameters while sharing a probabilistic density. Since there are $(t-2)$ potential change-points in a range, the most plausible change-point ($\tau^*$) should be chosen to obtain the log-likelihood ($LL_{\tau^*}$). To do so, for every candidate change-point ($\tau$) from 2 to $(t-1)$, two maximum log-likelihoods are computed for two sub-ranges [see Eq. (2)], and then an optimal change-point ($\tau^*$) is chosen such that the sum of the two log-likelihoods could be maximized [see Eq. (3)].

$$LL_\tau = \max_{\theta_L}[\log(f_{\theta_L}(x_1)) + \cdots + \log(f_{\theta_L}(x_\tau))] + \max_{\theta_R}[\log(f_{\theta_R}(x_{\tau+1}) + \cdots + \log(f_{\theta_R}(x_t))] \qquad (2)$$

In Eq. (2), $\theta_L$ and $\theta_R$ are different parameter sets of a probabilistic density function for the left and right ranges separated by a change-point ($\tau$). The statistic ($z^*$) that is used to reject the null hypothesis is the difference between the log-likelihood ($LL_0$) under the null hypothesis and that ($LL_{\tau^*}$) under the alternative hypothesis with the optimal change-point ($\tau^*$). The test statistic represents the ratio between two different likelihoods [see Eqs. (3) and (4)].

$$z^* = LL_{\tau^*} - LL_0 \qquad (3)$$
$$\tau^* = \underset{\tau}{argmax}\, LL_\tau \qquad (4)$$

Unfortunately, there is no typical probabilistic density function from which a statistic is sampled under the assumption that the null hypothesis is true. Setting up an empirical distribution is necessary to obtain a threshold to reject the null hypothesis. In principle, the full permutations of data in a given sequence should be enumerated to derive the distribution. The computing time for this, however, is exorbitant. As a practical solution, a sampling scheme was employed in the original paper [18]. A random sample of an arbitrarily chosen size was extracted from the original sequence and regarded as the left sub-sequence. The remainders of the variables were then regarded as the right sub-sequence. This scheme was appropriate because variables in different time intervals were assumed to be independent [18]. A test statistic was computed from the two sub-sequences using Eqs. (2) and (3). Once this sampling task was sufficiently repeated, an empirical distribution of the test statistic could be obtained. If an optimal test statistic ($z^*$) ranks in the top 5% of the distribution, the null hypothesis can be rejected, and the corresponding change-point is accepted. This statistical test can confirm the change-point found by the algorithm for every time window. In the present study, a 5% significance level was chosen because it is a market standard, and the sampling task was repeated 100 times to derive an empirical distribution for every time window.

The change-point detection algorithm was applied to features extracted from video images. The algorithm adopted both Gaussian and Dirichlet distributions. For each distribution density function, three different feature dimensions ($d$ = 3, 5, and 10) were tested. A Gaussian probabilistic density function has three groups of parameters: means, variances, and covariances. The number of parameters increases proportional to the feature dimension [= $d(d + 3)/2$]. The parameters are estimated every time a log-likelihood is computed. Eq. (5) denotes the multivariate Gaussian density function.

$$f(\pmb{x}_i) = \frac{1}{(2\pi)^{d/2}|\Sigma|^{1/2}} exp\left(-\frac{1}{2}(\pmb{x}_i - \mu)^T \Sigma^{-1}(\pmb{x}_i - \mu)\right) \quad (5)$$

In Eq. (5), $\pmb{x}_i = (x_{i1}, \dots, x_{id})^T$ is a feature vector in time step $i$, $\mu \in R^d$ is a vector that contains the mean of each feature component, and $\Sigma$ is a variance and covariance matrix for feature components.

A Dirichlet distribution density function is the best fit when dealing with compositional data. Features for the distribution are constrained such that their components should sum to 1. The support for a Dirichlet distribution of the order of $d$ is the ($d$-1)-simplex ($x_{ik} > 0$, $\sum_{k=1}^{d} x_{ik} = 1$). In other words, each point of a sequence should lie on the ($d$-1)-simplex. The number of parameters to be estimated equals the feature dimension, which are fewer than that for a Gaussian distribution density. The computing time for evaluation when using a Dirichlet distribution, on the other hand, is . Eq. (6) denotes the probabilistic density function for a Dirichlet distribution, where $\pmb{\alpha} = (\alpha_1, \dots, \alpha_d)^T$ is a set of parameters to be estimated.

$$f(\pmb{x}_i) = \frac{1}{B(\pmb{\alpha})} \prod_{k=1}^{d} x_{ik}^{\alpha_k - 1} \quad (6)$$

In Eq. 6, $x_{ik} > 0$, $\sum_{k=1}^{d} x_{ik} = 1$, and $B(\pmb{\alpha}) = \frac{\prod_{k=1}^{d} \Gamma(\alpha_k)}{\Gamma(\sum_{k=1}^{d} \alpha_k)}$.

It should be noted that each data point must be standardized and normalized prior to applying a Dirichlet distribution to detecting a change-point. Data are shifted with the global mean and scaled by the global standard deviation, and then normalized with an Expit function. An Expit function is the inverse of a multinomial logit function that transforms a point in $d$-dimensional space into a point on the $d$-simplex [see Eq. (7)]. The original paper [18] verified how the transformation of data do no harm to the statistical test established above.

$$g(\pmb{x}_i) = \left[\frac{e^{x_{i1}}}{1+\sum_k^d e^{x_{ik}}}, \dots, \frac{e^{x_{id}}}{1+\sum_k^d e^{x_{ik}}}, \frac{1}{1+\sum_k^d e^{x_{ik}}}\right]^T \quad (7)$$

### III. TESTBED AND DATA COLLECTION

The purpose of detecting change-points is to increase the measurement accuracy of multi-regime analysis for traffic surveillance. Computer vision-based traffic surveillance is largely affected by ambient conditions. Typically, video images vary significantly according to the time of day. Multi-regime analysis can be accomplished by identifying change-points online if model parameters are separately prepared for each recurrent regime. The testbed was chosen in an effort to reflect real-world conditions. An intersection approach located in Buchon city, Korea was selected, as shown in Fig. 2, and video images were taken for two days. The video frames taken in the first day were used to train an Autoencoder to reduce the feature dimensions.

The video frames of the second day were divided into times of day based on manual inspection. As a result, 7 different periods were considered: dawn, morning, daytime, late afternoon 1, late afternoon 2, evening, and nighttime. The criterion used to divide the late afternoon periods was based on shadow patterns at the site. Representative images for each period are shown in Fig. 3 along with threshold times. The next section will describe how the proposed algorithm was used to detect these thresholds in time and will feature the accuracy of the detection.

For models used to measure traffic density, 1,100 video images were randomly chosen from each period of the second day. For short periods from which the predefined number of images could not be chosen, video frames taken for the same period in both dates were added. For each period, 100 images were reserved for testing, and the remaining images were used for training. The overall model was trained and tested on the total amount of data (7,000 images for training and 700 images for testing).

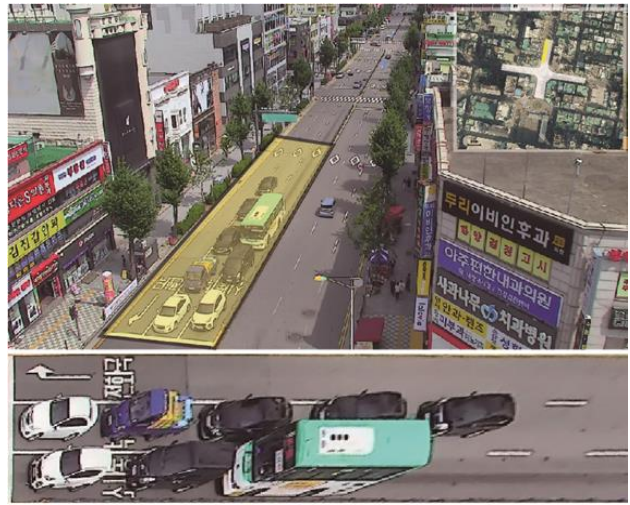

**FIGURE 2.** Testbed located in Buchon city, Korea .

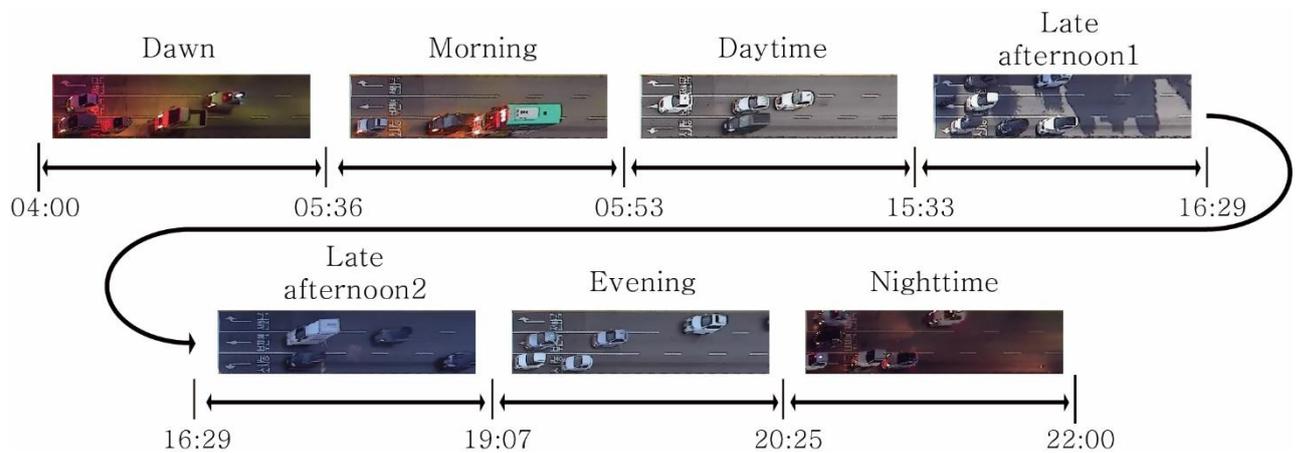

**FIGURE 3.** Different ambient conditions by times of day in the testbed

## IV. ANALYSIS RESULTS

### A. Reducing The Dimension Of Images

Training an Autoencoder is relatively easy by comparison with training other types of supervised deep learning models because it requires no human effort to tag images with labels. There is no need to secure labeled images because images are simultaneously used as both input and target when training an Autoencoder. That is, the mean squared error (MSE) is minimized between input images and the estimated output images. This simple scheme guarantees convergence when training an Autoencoder. A trained model is utilized to reduce the dimension of 128×512×3 mages into 3-, 5-, and 10-dimensional vectors.

Fig. 4 depicts a profile of reduced features extracted by an Autoencoder from video images. Each color indicates a specific feature extracted by an Autoencoder. It is impressive that the profile can show different patterns according to specific times of the day. The vertical lines in each graph indicate the predicted change-points, whereas small arrows in the top time line mark the ground-truth change-points that were determined by manually examining the video images. In cases where the Gaussian density was adopted for 3-dimensional abstracted features, changing patterns in the features are exactly consistent with those from the ground-truth change-points [see Fig. 4 (a)].

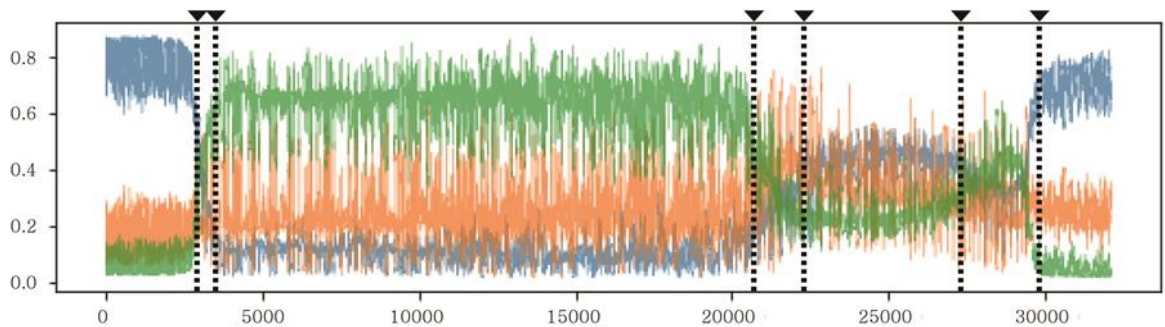

(a) Gaussian distribution for 3-dimensional features with the window size of 6,000

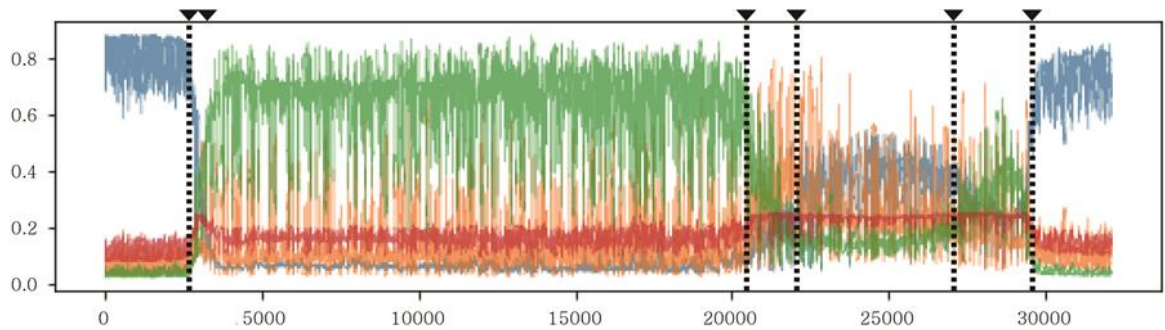

(b) Dirichlet distribution for 3-dimensional features with the window size of 5,000

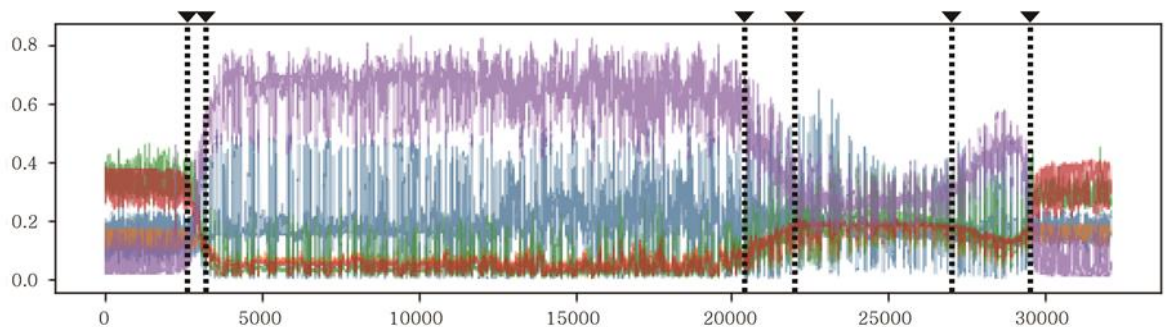

(c) Gaussian distribution for 5-dimensional features with the window size of 6,000

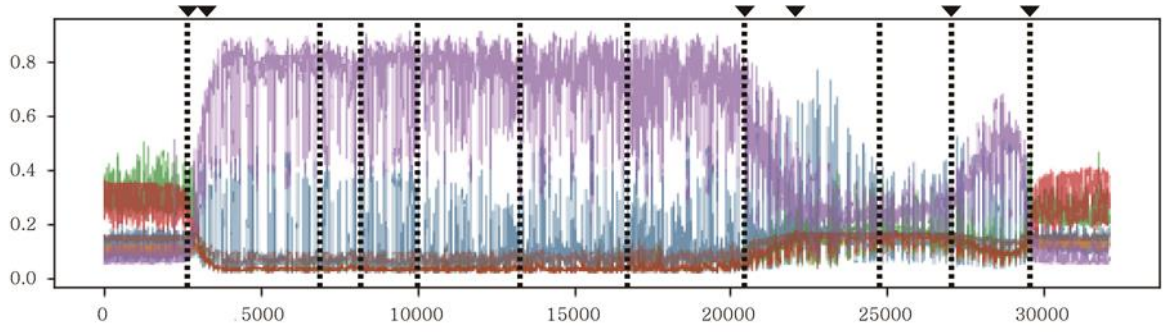

(d) Dirichlet distribution for 5-dimensional features with the window size of 5,000

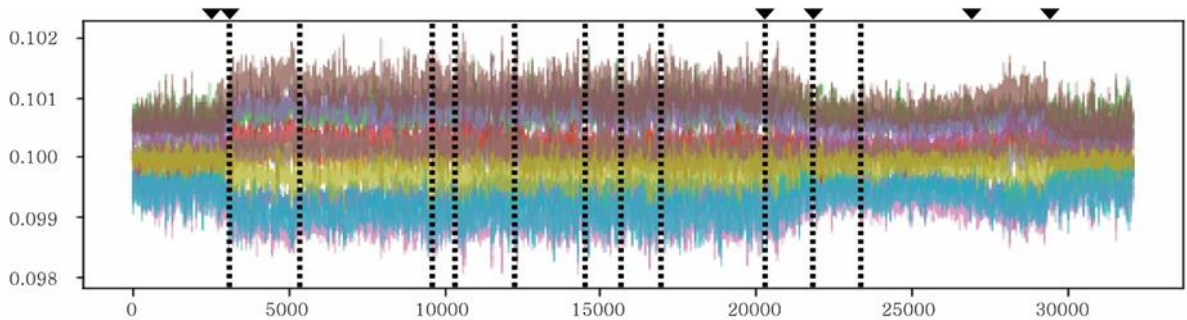

(e) Gaussian distribution for 10-dimensional features with the window size of 5,000

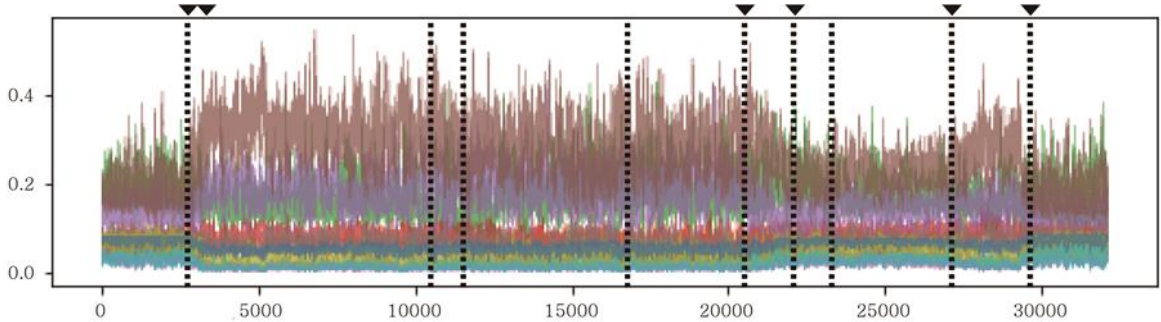

(f) Dirichlet distribution for 10-dimensional features with the window size of 5,000

FIGURE 4. The profile of reduced features for 128×512×3 mages

### B. Performance Of The Proposed Change-Point Detection Algorithm

The proposed change-point detection algorithm was applied to an image sequence of 22 hours (04:00-22:00) of the second day. The sequence used intervals of 2.0 seconds, and thus the total number of periods is 32,123. The algorithm employed the two probabilistic densities with 3 different feature dimensions. For each density function, 4 different window sizes were tested. The precision and recall of the change-point detection algorithm were computed for the total 24 different cases. When a predicted change-point approximated the ground truth within the predefined margin of error (= 300 intervals), the detection was regarded as successful.

Table 1 shows the test results of detecting change-points for the feature sequence of images. When using Gaussian density, a window size of 6,000 recorded the best performance with the two smaller feature dimensions (= 3 and 5). That is, both performance indices were 1.0, which meant that every ground truth change-point was detected with neither false positives nor false negatives. We also confirmed that a quite large window size (= 6000) is necessary for the perfect detection of change-points. It is interesting that the detection failed when the number of features was increased to 10 [see Fig. 4 (e)]. This implies that the use of many features to accommodate more conditions in an image may be ineffective in finding change-points attributable only to the time of day. More concretely, some of the 10 features may contain traffic conditions that can impede the distinguishing of ambient conditions. This result

8 | 11

proved that using 3 and 5 abstracted features is sufficient to extract necessary information from an image when using a Gaussian probabilistic density.

Regarding the Dirichlet density, the change-point detection algorithm was implemented after using the Expit function [Eq. (7)] to transform abstracted features into compositional data. Detection was never perfect when using Dirichlet density. The best performance was recorded when 3 abstracted features were used with a window size of 5,000. The precision was 1.0, but the recall was 0.833, since only 5 out of 6 ground truth change-points were properly detected and the remaining one was missed. Fortunately, there was no false positive. In terms of the results above, the Gaussian distribution was more robust than the Dirichlet in finding change-points for the times of day.

TABLE I
PRECISION AND RECALL WHEN DETECTING CHANGE-POINTS

(a) Test results from 3-dimensional features

| Gaussian distribution | Window size | Precision | Recall | Dirichlet distribution | Window size | Precision | Recall |
|---|---|---|---|---|---|---|---|
| | 3,000 | 0.417 | 0.833 | | 3,000 | 0.5 | 0.833 |
| | 4,000 | 0.375 | 0.5 | | 4,000 | 0.462 | 1 |
| | 5,000 | 0.75 | 1 | | 5,000 | 1.0 | 0.833 |
| | 6,000 | 1.0 | 1.0 | | 6,000 | 0.833 | 0.833 |

(b) Test results from 5-dimensional features

| Gaussian distribution | Window size | Precision | Recall | Dirichlet distribution | Window size | Precision | Recall |
|---|---|---|---|---|---|---|---|
| | 3,000 | 0.667 | 1.0 | | 3,000 | 0.235 | 0.667 |
| | 4,000 | 0.714 | 0.833 | | 4,000 | 0.308 | 0.667 |
| | 5,000 | 0.833 | 0.833 | | 5,000 | 0.4 | 0.667 |
| | 6,000 | 1.0 | 1.0 | | 6,000 | 0.2 | 0.333 |

(c) Test results from 10-dimensional features

| Gaussian distribution | Window size | Precision | Recall | Dirichlet distribution | Window size | Precision | Recall |
|---|---|---|---|---|---|---|---|
| | 3,000 | 0.222 | 0.667 | | 3,000 | 0.263 | 0.833 |
| | 4,000 | 0.154 | 0.333 | | 4,000 | 0.286 | 0.667 |
| | 5,000 | 0.273 | 0.5 | | 5,000 | 0.556 | 0.833 |
| | 6,000 | 0.111 | 0.167 | | 6,000 | 0.4 | 0.667 |

### C. Performance Of The Multi-Regime Approach In Measuring Traffic Density

Our previous work [30] developed a convolutional neural network (CNN) to count the number of vehicles in the image of a road segment. The CNN took the form of a regression model that could collectively count vehicles rather than adopting an object-detection model. The latter model is more accurate, since individual vehicles in an image can be detected and tracked. However, such accuracy might be redundant for traffic surveillance that can allow for some error.

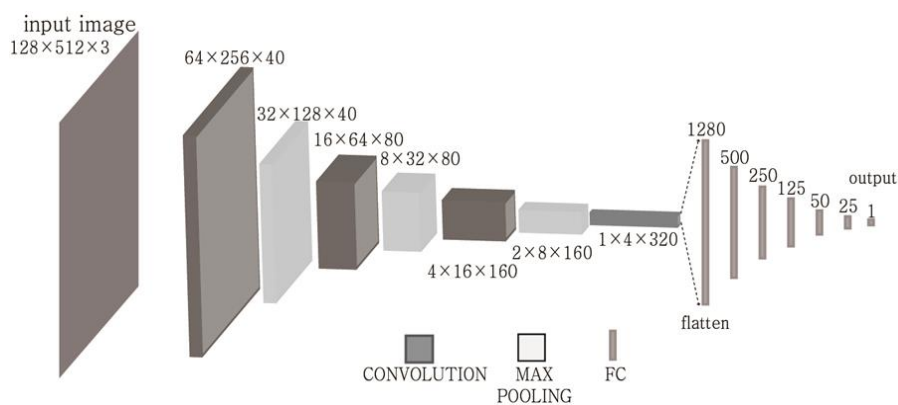

FIGURE 5. The CNN architecture used to predict the number of vehicles in an image.

The architecture of the regression-based CNN model is simple. Former convolutional layers of the model abstract features from an image, and the subsequent dense layers feed a final single node to collectively count the number of vehicles in the image. Labeling images for the model is also much easier than "state-of-the-art" detection models that require drawing a bounding box for every object in an image. Only the number of vehicles in an image can be counted to provide the present CNN model with labels.

An abbreviated CNN was devised by removing some of the convolutional layers from the regression-based CNN developed in our previous study [30]. This version of a CNN was applied to counting vehicles for each period that the change-point detection algorithm divided. Each period shared the same model architecture but had a separate set of model parameters, because the model was trained and tested on a separate dataset for each period. Fig. 5 shows the proposed architecture of a CNN to count vehicles. The model performance was measured by the root mean square error (RMSE) between observed and predicted vehicle counts.

The proposed model to count vehicles was also trained and tested using the entire dataset for all time periods. The test performance was measured for each period. The results are listed in Table 2. For the 5 periods highlighted in the table, separated learning outperformed integrated learning. The remaining two periods were too short to secure a sufficient amount of different images, although data collected during the two days were used. The performance of the overall model was slightly better than that of the separated model for the remaining two periods. The RMSE was computed for the integrated dataset for all periods based on both the separated and the integrated analyses (see Table 3). As a consequence, multi-regime analysis outperformed integrated analysis.

TABLE 2
MODEL PERFORMANCE USED TO MEASURE TRAFFIC DENSITY (RMSE) FOR DIFFERENT TIMES OF THE DAY

| RMSE | Dawn (04:00-05:36) 96 min. | Morning (05:36-05:53) 17 min. | Daytime (05:53-15:33) 580 min. | Late afternoon1 (15:33-16:29) 56 min. | Late afternoon2 (16:29-19:07) 158 min. | Evening (19:07-20:25) 78 min. | Nighttime (20:25-22:00) 95 min. |
|---|---|---|---|---|---|---|---|
| Separate training | 0.345 | 0.917 | 0.895 | 1.184 | 1.150 | 0.712 | 0.615 |
| Integrated training | 0.467 | 0.904 | 1.464 | 0.880 | 1.346 | 0.742 | 0.792 |

TABLE 3
MODEL PERFORMANCE USED TO MEASURE TRAFFIC DENSITY (RMSE) FOR ALL TIME PERIODS

| Separate training | 0.863 |
|---|---|
| Integrated training | 1.211 |

**V. Conclusions**

The present study dealt with the problem whereby computer vision-based traffic surveillance technology must account for the effects of different times of the day. A multi-regime analysis was set up and tested for counting vehicles in an image. The adopted change-point detection algorithm was robust in finding changing patterns in images according to times of the day. A statistical test method supports the theoretical background that was used to determine thresholds that separate multiple periods. An Autoencoder was used to extract a small number of features from an image containing ambient conditions. Such an abstraction made it possible to implement the change-point detection algorithm within a practical computing time.

The test results from counting vehicles in the testbed verified that the proposed multi-regime analysis outperformed an integrated model that was trained on the entire dataset that encompassed all periods. The test results proved that the proposed change-point detection algorithm could be applied to real-world online traffic surveillance. In further studies, robust models for measuring traffic volumes and speeds will be tested together using the proposed change-point detection algorithm. For example, the present approach would be useful when used with "state-of-the-art" object detection models for traffic surveillance.